# Robust Classification with Sparse Representation Fusion on Diverse Data Subsets

Chun-Mei Feng[a]; Yong Xu[a,b*]; Zuoyong Li[c]; Jian Yang[d]

[a] *Bio-Computing Research Center, Harbin Institute of Technology, Shenzhen 518055, Guangdong, P.R. China*

[b] *Key Laboratory of Network Oriented Intelligent Computation, Shenzhen 518055, Guangdong, P.R. China*

[c] *Fujian Provincial Key Laboratory of Information Processing and Intelligent Control (Minjiang University), Fuzhou 350121, China*

[d] *School of Computer Science and Engineering, Nanjing University of Science and Technology, Nanjing 210094, Jiangsu, P.R. China*

**Abstract**

Sparse Representation (SR) techniques encode the test samples into a sparse linear combination of all training samples and then classify the test samples into the class with the minimum residual. The classification of SR techniques depends on the representation capability on the test samples. However, most of these models view the representation problem of the test samples as a deterministic problem, ignoring the uncertainty of the representation. The uncertainty is caused by two factors, random noise in the samples and the intrinsic randomness of the sample set, which means that if we capture a group of samples, the obtained set of samples will be different in different conditions. In this paper, we propose a novel method based upon Collaborative Representation that is a special instance of SR and has closed-form solution. It performs Sparse Representation Fusion based on the Diverse Subset of training samples (SRFDS), which reduces the impact of randomness of the sample set and enhances the robustness of classification results. The proposed method is suitable for multiple types of data and has no requirement on the pattern type of the tasks. In addition, SRFDS not only preserves a closed-form solution but also greatly improves the classification performance. Promising results on various datasets serve as the evidence of better performance of SRFDS than other SR-based methods. The Matlab code of SRFDS will be accessible at http://www.yongxu.org/lunwen.html.

[*] Corresponding author
*Email address:* yongxu@ymail.com (Yong Xu)

# 1 Introduction

Sparse Representation (SR) has attracted much attention owing to its high accuracy in pattern classification especially in classification of high-dimensional data such as images [1-4]. However, SR suffers from the difficulty of high computational cost. What's more, time-consuming methods are always unsuitable in practice [5, 41, 42]. In light of above issues, designing efficient methods with high accuracy is definitely crucial.

To obtain computationally efficient representation methods, researchers in the field of pattern recognition have made many efforts [6, 7]. Since the iterative optimization style of $l_1$ norm constraint in SR results is time consuming, norm based constraints of other types are proposed [8-12], of which one typical instance is the $l_2$ norm based constraint, termed as "hard sparse representation" method i.e. Collaborative Representation (CR). Because of its nature of closed-form solution, the computation time can be greatly reduced. As an $l_2$ norm based algorithm, CR has been used in many scenarios [12-14]. It still shows reasonably good accuracy though it is quite computationally efficient. However, experiments reveal that even though CR may outperform SR in some cases, CR may be dwarfed by SR under some special conditions, conversely. In other words, "sparseness" does have its value to improve the classification performance. Fortunately, "sparseness" can be established based on the idea of CR in a special way [5]. In particular, the proper establishment of "sparseness" on the basis of CR can gracefully inherit the advantages of both SR and CR, obtaining good classification accuracy and maintaining computational efficiency [5].

For an image classification task on deformable objects, we usually encounter the difficulty that the number of samples in a class is very limited, and the appearance of the object is changeable. For two-dimensional data, we can deal with it by using one dimension to represent instances, and another to represent the attributes (features). For example, breast cancer data includes a binary dependent variable (known as class labels) in addition to these predictors, indicating whether breast cancer exists. Since this data has a small number of measurable features, the number of instances is greater than that of features. As a consequence, how to make the classification method adaptive to the potential diversity of samples is the focus of study. In classification problems, SR firstly encodes the test samples into a sparse linear combination of all training samples, and then tries to seek the best classification scheme for the test samples by obtaining the minimum classification error. Some methods based on sample generation have been put forward and applied. For example, bagging extracts samples from the original samples to construct the training sets, of which one training set is used to construct one model. Yet in this framework, all the training sets are independent of one another, i.e. some samples may be selected for multiple times, while some others may never be selected [15, 40]. In addition, another type of methods utilizing virtual and reasonable face images have been proposed to improve the performance of face recognition [16]. In fact, we find that with the help of virtual face images, a sparse or collaborative representation algorithm usually has a higher recognition rate. A shortcoming of these methods is that these methods [16] are only applicable to face recognition problems rather than more general problems.

From the viewpoint of sampling, the set of all available samples are just a part of the data including observed and unobserved data. In other words, even if we have a large-scale dataset, it is still only a result of the sampling process. Moreover, no matter how big the available data we are facing, we need only to get a subset of it, though there are numerous possible data subsets. In this sense, if we treat the obtained data subset as a whole of the data, the essence and the possible numerous forms of the data will not be grasped.

In this paper, we propose a simple but competent method for robust classification by producing reasonable Diverse sample Subsets and conducting Sparse Representation Fusion of subsets (SRFDS). Our method outperforms the state-of-the-art algorithms in this area by the advantages as follows: (1) It constructs diverse subsets from available samples in a scientific manner. (2) It chooses the classifier CR, due to its nature of closed form solution, so as to obtain desirable classification results. (3) It demands no special requirement on the pattern type of the task, which makes it applicable for general problems. (4) It's only necessary to deal with the subsets of available samples generated by the sampling process instead of the whole dataset, which allows us to obtain various possible representations of the data distributions. The contribution of this work can be summarized as follows: (1) We develop a novel classification method, in which the training samples are replaced with randomly generated subsets and CR is adopted to classify data. Thus, traditional CR enjoys the advantage of various possible representations which are caused by data distributions. (2) This method allows the training set being fully learned. It not only has the advantages of closed form solution, simple model and faster speed but also greatly improves the classification accuracy. (3) Extensive experiments on image and two-dimensional data clearly show that SRFDS outperforms the state-of-the-art SR based methods. We run our method as well as six state-of-the-art programs in this area on image and two-dimensional datasets, which forcefully reveals the advantage of our method.

## 2 Background

### *2.1 Presentation of sparse representation*

Sparse representation is a class of methods to convert the probed data into the linear combination of measurement matrices. The probed data is termed as test samples and the measurement data is named as training samples. It can be expressed as:

$$\mathbf{y} = \mathbf{X}\mathbf{a}, \qquad (1)$$

or

$$\mathbf{y} = \mathbf{x}_1 a_1 + \mathbf{x}_2 a_2 + ... + \mathbf{x}_m a_m. \qquad (2)$$

Herein $\mathbf{X} = [\mathbf{x}_1, \mathbf{x}_2, ..., \mathbf{x}_m] \mathbb{R}^{d \times m}$ is the matrix representing the given data, where each $\mathbf{x}_i$ represents one sample, and $\mathbf{a} = [a_1, ..., a_m]^T$ is the vector composed of the reconstruction coefficients, where each $a_i$ represents the reconstruction coefficient for sample $\mathbf{x}_i$. $\mathbf{y} \in \mathbb{R}^d$ represents the set of test samples, in which $d$ is the number of dimensions and $m$ is the number of samples. However, it is difficult to get a unique solution to (1). It is the intensive nature of the coefficients that leads to the redundancy of solutions. Thus, the concept of sparsity is introduced to solve the obstacle of intensity of coefficients. The method of sparsification is to transform the set of the test samples (represented by a vector) into the linear combination of training samples (represented by a set of vectors) so that many zero values exist in the coefficients. The sparse solution can be acquired by solving the following optimization problem:

$$\hat{\mathbf{a}} = \arg\min \|\mathbf{a}\|_p \quad s.t. \ \mathbf{y} = \mathbf{X}\mathbf{a}, \qquad (3)$$

where $\|\ \|_p$ refers to the $l_p$-norm constraint and $p \in [0 \sim 1]$. Specially, $l_p$ can also be defined as $l_{2,1}$-norm. Many studies have been conducted to investigate the effect of $l_0$, $l_1$ and $l_{2,1}$-norm on performance of sparse representation [17-20], and at last it's proved that $l_p$-norm could get better result when $p = 0.1, 1/3, 0.5$ and 0.9 [21, 22]. For classification task, $\mathbf{y} \in \mathbb{R}^d$ can be reconstructed by a linear combination of $\mathbf{X}$. The representation residual $r_i(\mathbf{y})$ can be obtained by follows:

$$r_i(\mathbf{y}) = \left\| \mathbf{y} - \mathbf{X}_i \hat{a}_i \right\|_2^2, \quad \forall_i \in \{1, 2, ..., c\}. \tag{4}$$

Then, we classify the test samples by finding which class has the smallest representation residual for the sample $\mathbf{y}_i$, which is formulated by $C(\mathbf{y}) = \arg\min_i r_i(\mathbf{y})$. To make the solution be obtained easily, the Lagrange multiplier is introduced in the model. Thus, the sparse model can be reformulated as:

$$\hat{\mathbf{a}} = L(\mathbf{a}, \lambda) = \arg\min \|\mathbf{y} - \mathbf{X}\mathbf{a}\|_2^2 + \lambda \|\mathbf{a}\|_p, \tag{5}$$

where $\lambda$ is the Lagrangian multiplier. The optimal solution to (5) can be obtained efficiently. If $p = 2$, this model can be termed as Collaborative Representation (CR). In this condition, this problem belongs to the convex optimization problem possessing a closed form solution, which greatly reduces the computational cost. Although the model has no explicit sparse constraint, it leads to the mutual exclusion among different training samples from all classes. All samples with correct class labels has the potential to induce larger $l_2$ norm values and smaller reconstruction errors. It can be formulated by follows:

$$\hat{\mathbf{a}} = \mathbf{P}\mathbf{y}, \tag{6}$$

where the projector $\mathbf{P} = \left( \mathbf{X}^T \mathbf{X} + \lambda \cdot \mathbf{I} \right)^{-1} \mathbf{X}^T$ is independent of $\mathbf{y}$. The biggest benefit of (6) is that the projector $\mathbf{P}$ can be obtained in advance based on the training samples. For brevity, the afterward steps of optimizing each test sample are omitted. At last, we try to get the value of the normalized residual for each class by solving

$$r_i(\mathbf{y}) = \left\| \mathbf{y} - \mathbf{X}_i \hat{a}_i \right\|_2^2 \bigg/ \left\| \hat{a}_i \right\|_2^2, \quad \forall_i \in \{1, 2, ..., c\}, \tag{7}$$

and then we classify $\mathbf{y}$ by labelling each $\mathbf{y}_i$ with the sample having the smallest representation residual $C(\mathbf{y}) = \arg\min_i r_i(\mathbf{y})$.

## 2.2 Classification of techniques on sparse representation

SR is a sparse form of Compressed Sensing (CS), which was widely used in signal processing [23]. Its deformation methods have been applied to image classification and have achieved notable performance [24-26]. SR techniques can be briefly divided into the following four categories:

(1) *Greedy strategy approximation*: SR with greedy strategy solves the NP-hard problem of minimizing the $l_0$-norm. This method searches for the optimal solution in each iteration, and finally arrives at an approximate solution. [23, 27, 28].

(2) *Constrained optimization*: SR with sparse constraints are proposed to meet the requirement of sparsity. Furthermore, the problem of $l_0$-norm minimization has been solved using this technique [23, 29]. Typically, the Lasso regularization using $l_1$ norm is widely applied in this area, yet $l_p\ (0<p<1)$ and $l_{2,1}$-norm constraint are the most representative ones [10, 30].

(3) *Proximity algorithm*: Proximal algorithm effectively solves the problem of non-smoothness, constraint and large scale in traditional optimization methods [31]. In these algorithms, at first the original sparse problem is transformed into a specific model. Then, the proximity algorithm is used to solve this model [32].

(4) *Homotopy algorithm*: Based on related techniques in topology, homotopy algorithms keep track of the path of all the solutions as well as the parameter changes during iteration, which has been successfully applied to the $l_1$-norm minimization problem [23, 33].

## 3 Methodology

### 3.1 Motivations and potential drawbacks of traditional sparse representation

Though SR based techniques are widely applied in many fields because of its high accuracy, it suffers from two disadvantages as follows: (1) Traditional SR based algorithms possess low speed and high computational cost. Although some techniques, e.g. virtual face images, have been proposed to improve the efficiency of SR on face recognition, these techniques haven't been spread to other areas. (2) Traditional SR based algorithms can learn their models from a given training set in only one viewpoint, hence the knowledge is limited and biased. They underexploited the training samples, since they ignore the uncertainty hidden in the training samples. In traditional SR based algorithms, many latent hidden features of the train samples will be lost, especially for the dataset with large sample sizes.

To overcome those obstacles above, the method of CR with closed-form solution is proposed. With this method, we try to solve those problems, respectively. (1) Classifier CR replaces the sparse constraint used in SR with the $l_2$-norm constraint, and adopts the $l_2$-norm constraint from SR. Since the convex optimization problem in this new model can be solved by a closed-form solution, it reduces the computational cost dramatically. (2) We proposed a new method SRFDS to exploit the latent hidden features training samples and obtain various possible data representations. Our classification method can be viewed as an improvement over CR. When the method is integrated with the generated multiple subsets, it indeed models the performance of the classification algorithm on different sample aggregations. Because the results of the algorithm on all data subsets are summarized, the final decision is robust and accurate. The random way of generating data subsets eliminates the uncertainty of the representation and makes the result representative. And different representations of classification can be obtained by the sample subset in data.

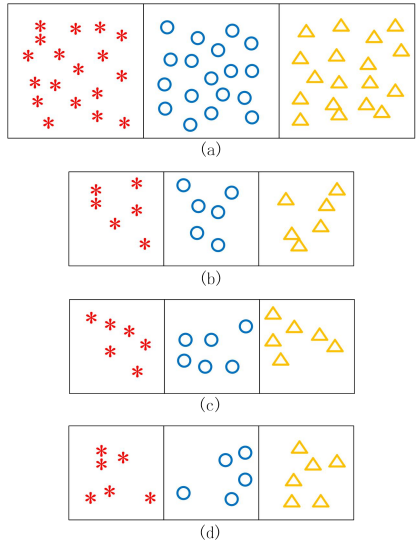

Figure 1. Various possible representations of the data distributions. (a) represents the training samples of three different classes. (b), (c) and (d) are the three subsets representing the multiple distributions of the data. The distribution structure of data in subsets makes the structure of a subset of samples clearer so as to improve the classification accuracy.

For example, Figure 1 displays the various possible representations of the data distributions. (a) is the given training set, and it's assumed that there are three classes, each containing 18 samples. The three subsets representing multiple distributions of the data are described in (b) (c) (d), respectively. Herein each subset is independent of each other. As shown in Figure 1, the distributions of data clearly show the inner structure of a few samples. The diverse structures in the subsets are also mined so that the original training set can be learned from various aspects. With this method, the classification accuracy is improved. In other words, the expression level and contribution to the classification of each training sample in different subsets are also different. As seen in Figure 2, suppose (e), (f) and (g) represent three different training subsets respectively. Figure 2 shows that the contribution to classification and expression of each training sample in these subsets are various. For example, the expression of training_sample 1 in (f) is the lowest, but this does not affect its large contribution in the subset (g). The different contributions of training samples in each subset express the multiple distributions existing in the training set perfectly. Our goal is to subdivide a given training sample to obtain various possible data distributions. Therefore, SRFDS is a universal method and is not limited to the pattern type of training data.

## 3.2 Description of the Proposed Method

We propose a simple method to obtain the numerous forms of the training samples. Our goal is to generate multiple sample subsets on the given samples, which allows us to obtain a number of possible representations of the data. In order to improve the accuracy, the classification algorithm is deployed on resultant different sample subsets. A summary of the results of all data subsets makes the final decision more robust and accurate. The

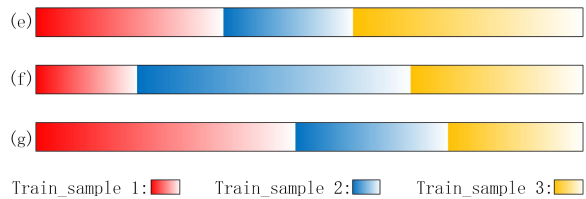

Figure 2. The contribution of each training sample to the classification of training samples in different subsets.

idea of the proposed method is summarized in Figure 3, and the details of the steps are given as follows:

**Step 1:** For a given training set $\mathbf{X} = \{\mathbf{X}_1, \mathbf{X}_2, ..., \mathbf{X}_c\} \in \mathbb{R}^{d \times m}$ and test set $\mathbf{y}$, we generate a number of sub-training sets $sub_{(i)}$ ( by default, four subsets ) randomly. $\mathbf{X}_i$ refers to the matrix representing the $n_i$ train samples of the $i$-th class. Assuming that each class has $N$ training samples, the total number of all training samples is $m = cN$, where $c$ is the number of categories. We firstly randomly extract half of all the training samples and store them in subset $sub_{(1)}$ of size $cN_1$ and the rest in subset $sub_{(2)}$ of size $cN_2$, where $N = N_1 + N_2$. Then the results of the second extracts are returned and stored in subset $sub_{(3)}$ of size $cN_1$ and subset $sub_{(4)}$ of size $cN_2$ respectively. After doing the above operation on the training samples from each class, the complete four subsets are obtained.

**Step 2:** Based on the sub-training sets obtained in **step 1**, we get the projection $\mathbf{P}_{(i)}$ and the reconstruction coefficients $\mathbf{a}_{(i)}$ by (6), where $i = 4$. The test set $\mathbf{y} \in \mathbb{R}^m$ is reconstructed by the linear combination of training subsets. Then, for each $i$ and $j$, we calculate their reconstruction error of the $j$-th class of the $i$-th subset and represent it as $e_{(i,j)}^{st}$. With this method, we complete the modeling of classification algorithms on different sample aggregations.

**Step 3:** Finally, we get the total reconstruction error based on $e_{(i,j)}^{st}$ and then perform reclassification. We fuse these reconstruction errors to obtain the fused classification error. The fused classification error denoted by $e_z^f$, where $z$ represent the $z$-th time to construct training subset.

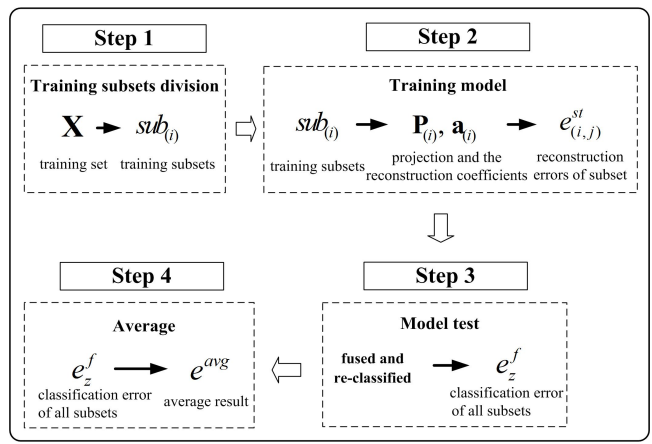

Figure 3. Flow-chart for the proposed method, which describes the classification process with diverse subsets.

**Step 4:** Since the training subset is randomly constructed each time, we take the average value over all calculated classification errors $e^{avg} = (e_1^f + e_2^f + ... + e_z^f)/z$ as the final classification error on the given training set after steps 1, 2, and 3 are repeated ten times.

The detailed procedure of the proposed method is presented in Algorithm 1.

| **Algorithm 1.** The pseudocode of SRFDS |
|---|
| **Input:** $\lambda$, $\mathbf{y}$ and $\mathbf{X}$, |
| **Output:** $\mathbf{P}_{(i)}$, $\mathbf{a}_{(i)}$, $e_{(i,j)}^{st}$, $e_z^f$ and $e^{avg}$. |
| 1. Sub-training sets $sub_{(i)}$ are calculated by **Step 1**; |
| 2. $\mathbf{P}_i$ is obtained via $\mathbf{P} = (\mathbf{X}^T\mathbf{X} + \lambda \cdot \mathbf{I})^{-1}\mathbf{X}^T$, where $\mathbf{X}$ is substitute by $sub_{(i)}$; |
| 3. $\mathbf{a}_{(i)}$ is calculated by (6); |
| 4. Reconstruction errors of each subsets $e_{(i,j)}^{st}$ are calculated by **Step 2**; |
| 5. Final classification error $e_z^f$ is calculated by **Step 3**; |
| 6. Average value $e^{avg}$ of classification errors is calculated by **Step 4**. |

## 4 Experiments and results

The experiments mainly test the effect of the proposed method on classification task, because the initial goal of our method is to investigate the training data comprehensively. To test the universality of our method, we conducted classification experiments by running our method as well as the state-of-the-art SR based methods, such as CR [34], INNC [35], Homotopy [36], FISTA [6], PALM [37], and FCM [38] on two images and two two-dimensional datasets. These methods include $l_1$-norm constrained SR method, $l_2$-norm constrained SR method, classical homotopy SR method, fast shrinkage SR method, augmented Lagrangian SR method, combinations of SR and other methods and fusion classification method. Since these methods are typical sparse representation algorithms and have achieved good performances in classification, the effectiveness of our proposed method can be reasonably proved by outperforming these mainstream algorithms. Content of this section is organized as follows. Firstly, the experimental data are described in section 4.1. Secondly, the baseline methods are briefly described in section 4.2. The time cost of subset partition is recorded to analyze the efficiency of our method in section 4.3. Finally, the universality of our method is verified by classification experiments on different types of data, such as images and two-dimensional multivariate data.

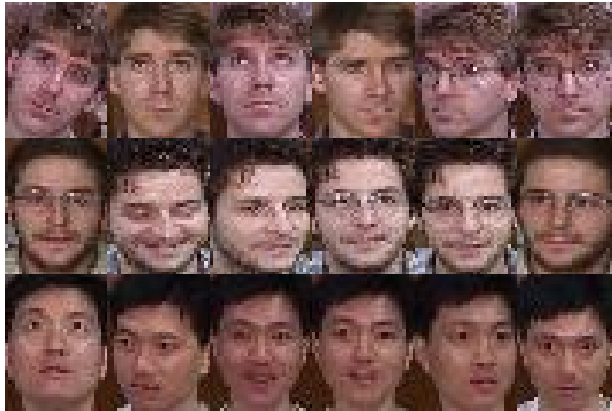

Figure 4. Sample images in the GT face dataset. Each row displays different images of different subjects.

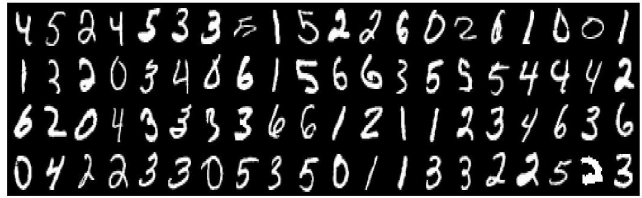

Figure 5. Sample images in the MNIST database.

*4.1 Datasets*

Two types of datasets image and two-dimensional multivariate data are used in the experiments, of which the image datasets come from Georgia Tech (GT) Face database and MNIST handwritten digit images database. The number of samples for each class in the image data is limited, but the appearance of the objects are easy to change. Firstly, for GT face data, uncertainties such as illumination, pose, occlusion and angle all have important effect on the classification effect. The 750 face images in GT face dataset originate from 50 persons of whom each has 15 face images [39]. These images focus on different facial expressions as well as different inclinations and external lighting conditions. All images in GT face data are resized to 30 × 40. Some sample images in GT face dataset are presented in Figure 4. Before training and testing, each image is converted into a grayscale image. Secondly, the MNIST handwritten data can be categorized in ten classes. The images vary in writing styles and shapes by different annotator. It consists of 6000 handwritten digits and equally distributed in 10 classes. All images in this dataset are resized to 28 × 28. Some images in MNIST dataset are shown in Figure 5. Each row displays different images of different digits.

In addition, we also use the two-dimensional multivariate datasets to conduct the experiments. Diagnostic Wisconsin Breast Cancer (DWBC) dataset consist of a matrix that includes 424 instance and 30 attributes. All instances are patients with breast disease, of which 212 are malignant and the rest are benign. Another dataset Clinical Coimbra Breast Cancer (CCBC) consists of a matrix that includes 104 instance and 8 attributes. These instances are divided into 52 patients and 52 healthy controls. These two-dimensional multivariate data possesses some properties of the anthropometric data and related predictors of breast cancer. If the several biomarkers for the exploration of causative factors of breast cancer, and these predictions are accurate, they will provide information can be obtained at http://archive.ics.uci.edu/ml/datasets.html. The aforementioned two kinds of data are completely different, so the universality of our method can be well tested.

### 4.2 Baseline methods

Six relevant methods are compared in our experiments to test whether the proposed method fulfills the classification task better. The compared methods in our experiment can be summarized as follows. (1) CR: CR is a SR method with $l_2$-norm constraints. The model is simple and fast, and possesses good performance in face recognition [34]. As a classifier of our method, CR can be regarded as a comparison method to verify the reliability of our method. (2) INNC (improvement to the nearest neighbor): This method combines the classification procedure of SR to improve the nearest neighbor classifier. It also achieves effective performance in face classification [35]. (3) Homotopy: Homotopy can be used to address the problem of nonlinear optimization effectively. By manually setting the initial value and gradually adjusting the homotopy parameters, the expected solution is obtained and the path of the complete solution is recorded [36]. (4) FISTA (fast iterative shrinkage-thresholding algorithm): FISTA is the proximity SR algorithm based optimization strategy. This method preserves the computational simplicity of iterative shrinkage-thresholding algorithm and has global convergence [6]. (5) PALM (primal augmented Lagrangian method): PALM belongs to SR method which is solved by Augmented Lagrange method (ALM) [37]. (6) FCM (fusion classification method): FCM is a classification method based on reconstruction error and normalized distance [38]. These SR based methods include $l_1$-norm and $l_2$-norm constraints as well as Homotopy and ALM, which are representative SR methods.

Table 1. Time cost of subset division.

| training set | GT | MNIST | DWBC | CCBC |
|---|---|---|---|---|
| no division | 0.3864(s.) | 26.7642(s.) | 0.0180(s.) | 0.0037(s.) |
| subset division | 0.4880(s.) | 27.1973(s.) | 0.0202(s.) | 0.0059(s.) |

### 4.3 Time cost analysis of subset division

Since we do not change the CR model, its complexity has not been changed, and the high speed of the closed-form solution is still retained. Our method only consumes extra time on the subsets of training data. Therefore, in the experiments, we record the time cost of training set division to test whether our method retains the original faster speed characteristic. Table 1 lists the time cost of subset division. As can be seen from this table, the extra time spent on dividing subset is relatively small. In other words, our method does retain the advantage of fast speed for closed-form solutions. And the advantage of our method in classification task is not at the expense of temporal resources.

### 4.4 Experimental settings and classification results on image datasets

The novel idea of the proposed method aims to study the training set comprehensively to obtain various possible representations of the data distributions. By generating multiple subsets on available samples, the classification accuracy can be improved. First, we conduct classification tests on image data. Table 2 and 3 show classification accuracies of different methods on the GT face and MNIST dataset. For GT face dataset, we extract first 6-14 images from each subject as the original training set, and the remaining images as testing set. For MNIST dataset, 6-16 images are extracted as the training set. And the parameter $\lambda$ is set to 0.1. For other comparative methods, the optimal value of parameter $\lambda$ for each image dataset is selected in the range of $\lambda = (10^{-6},...,10^2)$. As shown in Tables 2 and 3 that the proposed does obtain high classification accuracy in image data.

Table 2. Classification accuracies of different methods on the GT face database.

| Methods | Number of training samples per class | | | | |
|---|---|---|---|---|---|
| | 6 | 8 | 10 | 12 | 14 |
| SRFDS | **62.31** | **68.23** | **75.00** | **76.27** | **84.00** |
| CR | 55.56 | 59.43 | 64.00 | 68.00 | 72.00 |
| INNC | 56.00 | 61.71 | 63.60 | 70.00 | 74.00 |
| Homotopy | 59.33 | 65.71 | 66.80 | 72.67 | 74.00 |
| FISTA | 56.22 | 55.14 | 56.80 | 56.67 | 58.00 |
| PALM | 48.67 | 51.14 | 47.60 | 51.33 | 58.00 |
| FCM | 57.78 | 63.43 | 69.57 | 74.67 | 76.00 |

Table 3. Classification accuracies of different methods on the MNIST database.

| Mthods | Number of training samples per class | | | | | |
|---|---|---|---|---|---|---|
| | 6 | 8 | 10 | 12 | 14 | 16 |
| SRFDS | 68.09 | **69.34** | **71.10** | **72.18** | **74.10** | **74.06** |
| CR | **68.77** | 67.52 | 69.47 | 71.70 | 73.75 | 73.72 |
| INNC | 65.44 | 63.18 | 64.85 | 66.48 | 68.26 | 69.09 |
| Homotopy | 64.83 | 66.23 | 68.00 | 70.02 | 72.66 | 73.34 |
| FISTA | 67.01 | 68.96 | 70.78 | 71.30 | 73.86 | 73.33 |
| PALM | 63.01 | 64.44 | 65.56 | 66.21 | 67.42 | 67.33 |
| FCM | 68.20 | 65.76 | 67.31 | 69.15 | 70.68 | 71.32 |

*4.5 Experimental settings and classification results on two-dimensional multivariate datasets*

The diversity construction based on training set makes SRFDS more universal. It is necessary to test how well the various possible representations of the data distributions faciliate the classification task. Here, we test our method on two-dimensional multivariate data. Table 4 and 5 display classification accuracies of different methods on the DWBC and CCBC dataset. For DWBC dataset, the parameter $\lambda$ is set to 0.00001. And we set $\lambda = 0.001$ for CCBC dataset. It can be seen from these tables that SRFDS does obtain high classification accuracy in the two-dimensional multivariate datasets.

Table 4. Classification accuracies of different methods on DWBC dataset.

| Methods | Number of training samples per class | | | | | |
|---|---|---|---|---|---|---|
| | 30 | 32 | 34 | 36 | 38 | 40 |
| SRFDS | **94.81** | **95.14** | **95.14** | **94.77** | 95.03 | **95.58** |
| CR | 90.38 | 90.00 | 91.85 | 90.91 | 90.80 | 90.70 |
| INNC | 89.56 | 88.61 | 89.33 | 87.50 | 86.78 | 86.92 |
| Homotopy | 91.48 | 92.22 | 91.85 | 92.50 | 91.54 | 92.15 |
| FISTA | 69.51 | 70.28 | 63.20 | 68.48 | 76.44 | 84.88 |
| PALM | 92.03 | 92.22 | 92.13 | 90.62 | 91.09 | 90.99 |
| FCM | 93.13 | 93.06 | 93.82 | 94.60 | **95.11** | 94.48 |

Table 5. Classification accuracies of different methods on CCBC dataset.

| Methods | Number of training samples per class | | | | | |
|---|---|---|---|---|---|---|
| | 30 | 32 | 34 | 36 | 38 | 40 |
| SRFDS | 77.05 | **79.50** | **79.17** | 78.12 | **75.36** | **75.42** |
| CR | 75.00 | 75.00 | 75.00 | 75.00 | 67.86 | 66.67 |
| INNC | 69.64 | 69.64 | 68.75 | 75.00 | 72.95 | 67.78 |
| Homotopy | **77.27** | 75.00 | 75.00 | 75.00 | 71.25 | 75.00 |
| FISTA | 72.73 | 75.00 | 63.89 | 59.37 | 67.86 | 66.67 |
| PALM | 72.73 | 77.47 | 72.22 | **78.12** | 75.00 | 75.00 |
| FCM | 75.00 | 79.00 | 79.01 | 75.00 | 75.00 | 75.00 |

## 4.6 Analysis of Experimental Results

The classification results enable the following observations:

(1) On the whole, the accuracy of classification increases with the increase of the number of training samples. Sufficient training samples make the generated subsets more representative, especially on the image datasets. This is mainly because that the $c$ in the label of image data which in our experiment is relatively large. As the number of training set increases, the constructed subsets are more likely to reveal multiple classes of structures. Since the $c$ in label of the ordinary two-dimensional multivariate data in our experiment is relatively small, the results show that this trend is weaker. Especially on CCBC dataset, the noise exists in some samples of the training dataset leads to over-fitting and reduced the recognition rate. However, our method still has a lower error rate, indicating that the proposed method is more robust.

(2) Most of these methods do not guarantee consistent performance on different datasets. Generally, the method that works well on the image datasets cannot achieve similar effects on other datasets. For example, the performance of CR on the MNIST dataset is outstanding, but is far from satisfactory on the CCBC dataset. It shows that these methods are not universal and not suitable for various data.

According to these observations, we summarized the following results about the proposed method in this research: First, SRFDS is suitable for classification task. Second, results on various datasets show that the effectiveness of SRFDS in classification does not limit to the pattern type of the task. Because the diverse constructions of sample sets generate multiple subsets on available samples, the various possible representations of the data distributions can be obtained. And these diverse subsets promote the comprehensive learning of the original training set. Third, SRFDS method possesses improved classification accuracy and fast speed. Fourth, by using diverse subsets and representation fusion, SRFDS could intuitively reduce computation complexity by a large margin while directly using SR on large scale dataset might cause unaffordable computation cost.

## 5 Conclusion

In real world, the data depicting an object might be diverse and complex. In particular, samples of the data depicting a deformable object always vary. Though SR might has a high accuracy, it neglects the variability and uncertainty of samples. Herein we designs a simple and universal method SRFDS. The most prominent characteristic of this method is that it may seem to be suitable for the pattern types of different tasks. Different from the previous methods of changing the classifier itself for classification fusion, we diversify the training set for the same goal. Since the contributions of the training samples on different subsets are different, we obtain various possible data representations. These different data representations make the training set to be fully learned. Thus, the test results are more roust, which is very useful for complex problems. By generating multiple subsets on available samples and using CR as the classifier, SRFDS not only retains the closed form solution but also improve the classification accuracy. More importantly, SRFDS is simple, and outstanding classification performances are obtained at low computation cost. Besides, the experimental results on different kinds of data also confirm that our idea is reasonable and effective. A limitation of SRFDS is that the effect of the reconstruction error of each subset in the final evaluation step has not been explored. It is necessary to design a system to automatically set weight for each reconstruction error. In the future, we will attempt to address this problem so as to improve the robustness.

## Acknowledgment

This work is partially supported by Guangdong Province high-level personnel of special support program (No. 2016TX03X164), National Natural Science Foundation of China (61772254), and Fujian Provincial Leading Project (2017H0030).

## Author Biography

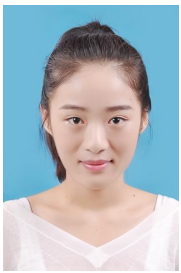

**Chunmei Feng** received her M.S. degree in School of Information Science and Engineering at Qufu Normal University in 2018. She is currently pursuing the Ph.D. degree in computer science and technology at Harbin Institute of Technology, Shenzhen, China. Her research interests include, pattern recognition, feature extraction and deep learning.

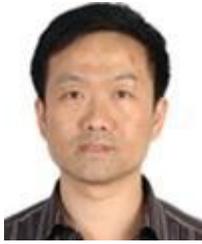
**Yong Xu** was born in Sichuan, China, in 1972. He received the Ph.D. degree in Pattern recognition and Intelligence System at the Nanjing University of Science and Technology (NUST) in 2005. Now, he works at Harbin Institute of Technology, Shenzhen, China. His current interests include pattern recognition, biometrics, machine learning and video analysis. More information please refer to http://www.yongxu.org/lunwen.html.

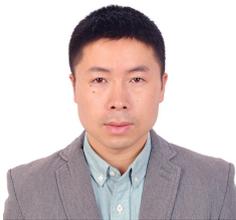
**Zuoyong Li** received the B.S. and M.S. degrees in Computer Science and Technology from Fuzhou University, Fuzhou, China, in 2002 and 2006, respectively. He received the Ph.D. degree from the School of Computer Science and Technology at Nanjing University of Science and Technology, Nanjing (NUST), China, in 2010. He is currently a professor in Department of Computer Science of Minjiang University, Fuzhou, China. He has published over 60 papers in international/national journals. His current research interest is image processing, pattern recognition and machine learning.

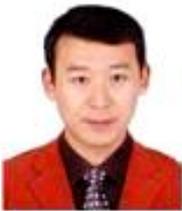
**Jian Yang** received the B.S. degree in Mathematics from Xuzhou Normal University, Xuzhou, China, in 1995, the M.S. degree in applied mathematics from Changsha Railway University, Changsha, China, in 1998, and the Ph.D. degree in the subject of pattern recognition and intelligence systems from the Nanjing University of Science and Technology (NUST), Nanjing, China, in 2002. In 2003, he was a Post-Doctoral Researcher with the University of Zaragoza, Zaragoza, Aragon, Spain. From 2004 to 2006, he was a Post-Doctoral Fellow with the Biometrics Center, Hong Kong Polytechnic University, Kowloon, Hong Kong. From 2006 to 2007, he was a Post-Doctoral Fellow with the Department of Computer Science, New Jersey Institute of Technology, Newark. Currently, he is a Professor with the School of Computer Science and Technology, NUST. He is the author of more than 50 scientific papers on pattern recognition and computer vision. His journal papers have been cited more than 1200 times on the ISI Web of Science, and 2000 times on Google Scholar.

His current research interests include pattern recognition, computer vision, and machine learning.